\documentclass{article} % For LaTeX2e
\usepackage{iclr2027_conference,times}

\usepackage{amsmath,amsfonts,bm}

\def\eqref#1{equation~\ref{#1}}
\def\1{\bm{1}}

\DeclareMathAlphabet{\mathsfit}{\encodingdefault}{\sfdefault}{m}{sl}
\SetMathAlphabet{\mathsfit}{bold}{\encodingdefault}{\sfdefault}{bx}{n}

\usepackage[hidelinks,hyperfootnotes=false]{hyperref}
\usepackage{url}
\usepackage{graphicx}
\usepackage{booktabs}
\usepackage{multirow}
\usepackage{placeins}
\usepackage{enumitem}
\usepackage{colortbl}
\usepackage{tabularx}
\usepackage[bottom]{footmisc}
\newcolumntype{Y}{>{\centering\arraybackslash}X}
\newcolumntype{L}{>{\raggedright\arraybackslash}X}
\definecolor{distancekvblue}{RGB}{232,242,252}
\title{Distance-KV: Exploiting Relative Distance for Efficient Long-Context Inference}

\author{
\textbf{Xianpeng Shang}\textsuperscript{1} \quad
\textbf{Canbin Huang}\textsuperscript{2,3} \quad
\textbf{Jiang Li}\textsuperscript{1} \quad
\textbf{Tian Lan}\textsuperscript{4} \\
\textbf{Qianyi Cai}\textsuperscript{5} \quad
\textbf{Xiaojun Quan}\textsuperscript{2}%
\thanks{Co-corresponding authors.} \quad
\textbf{Xiangdong Su}\textsuperscript{1}\footnotemark[1] \\[0.2em]
{\small
\textsuperscript{1}Inner Mongolia University \quad
\textsuperscript{2}Shenzhen Loop Area Institute \quad
\textsuperscript{3}Sun Yat-sen University
} \\
{\small
\textsuperscript{4}Kyoto University \quad
\textsuperscript{5}The Hong Kong University of Science and Technology (Guangzhou)
}
}

\iclrfinalcopy % Uncomment for camera-ready version, but NOT for submission.
\begin{document}

\maketitle
\lhead{Preprint}
\begin{abstract}
The memory usage and decoding latency of LLM inference grow rapidly with context length. To reduce these costs, key-value (KV) cache compression methods selectively retain cached states based on token importance or differences in attention patterns across heads. However, we discover that retrieval capability varies substantially with relative distance, even within the same attention head. To exploit this structure, we introduce Distance-KV, which learns a static KV retention pattern over the joint space of layers, attention heads, and relative distances. The pattern is learned offline with the language model frozen and reused across inputs to prune and compact the KV cache without online importance scoring. Across three backbone models and four long-context benchmarks, Distance-KV consistently achieves the best overall performance among competing KV cache compression methods, exceeding the strongest compression baseline by up to 9.3 points on RULER at 128K. On Llama-3.1-8B-Instruct at 128K, Distance-KV reduces KV cache memory by 65.4\% and achieves a \(1.66\times\) decoding speedup relative to Dense. Together, these results identify relative distance as an important structural dimension for understanding how LLMs retrieve information over long contexts and for designing more efficient inference methods.
\end{abstract}

\begin{figure}[b]
    \centering
    \includegraphics[width=\linewidth]{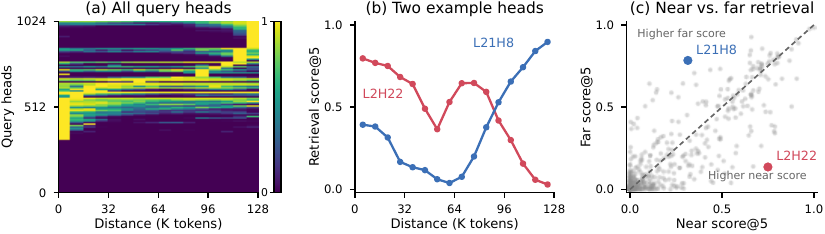}
    \caption{
\textbf{Retrieval capability varies across attention heads and distances.}
Results are shown for Llama-3.1-8B-Instruct.
\textbf{(a)} Per-head-normalized retrieval score@5 for all query heads, sorted by peak distance.
\textbf{(b)} Raw retrieval score@5 profiles of two example heads.
\textbf{(c)} Mean retrieval score@5 at near and far distances for every query head; the diagonal indicates equal scores.
}
\label{fig:motivation}
\end{figure}

\section{Introduction}

Extending the context length enables large language models (LLMs) to process long documents, code repositories, and multi-turn interactions, but the resulting growth of the key-value (KV) cache substantially increases memory use and decoding latency~\citep{comanici2025gemini25pushingfrontier,yang2025qwen251mtechnicalreport}. 
To reduce this overhead, prior work has explored KV retention at multiple granularities, including individual tokens, layers, and attention heads~\citep{zhang2023ho,cai2025pyramidkv,ge2024model}. 
In particular, work on retrieval heads identifies a small subset of attention heads as critical for retrieving relevant information from long contexts~\citep{wu2025retrieval}. 
Existing head-level methods assign each head a single score, label, or sparse attention rule, without explicitly modeling how its retrieval capability varies with relative distance~\citep{xiao2025duoattention,DBLP:journals/corr/abs-2406-14909}.

Figure~\ref{fig:motivation} reveals a limitation of this head-level characterization: an attention head's retrieval capability can vary markedly across relative distances (see Section~\ref{sec:distance-dependent-retrieval} for further analysis). 
Consequently, head-level allocation cannot distinguish between distance ranges where the same head exhibits strong and weak retrieval capability. 
As a result, retaining an entire head may consume cache budget at distances where it contributes little, whereas compressing it may discard useful states at other distances.
This within-head variation suggests that relative distance is not merely a positional coordinate but a key dimension for characterizing long-context retrieval.

Building on this observation, we propose Distance-KV, which learns a static KV retention pattern over the joint space of layers, query heads, and relative distances. 
This finer granularity allows the same query head to retain KV states over useful distance ranges while compressing the rest. 
For each model and cache budget, Distance-KV optimizes the pattern offline using synthetic retrieval examples while keeping the language model frozen.
The resulting pattern is independent of the content of both the context and the query and can therefore be reused across inputs and successive queries.
At inference time, the learned pattern is mapped to token positions and applied once to prune and compact the KV cache, requiring no online importance scoring or optimization. 

We evaluate Distance-KV on controlled retrieval, downstream long-context tasks, robustness across context lengths, and shared-context multi-turn reuse using three backbone models. Across these settings, Distance-KV consistently leads the competing compression methods and improves over the strongest compression baseline by up to 9.3 points on RULER at 128K. The reduction in retained KV states also translates into practical efficiency: on Llama-3.1-8B-Instruct at 128K, Distance-KV uses \(65.4\%\) less KV cache memory and achieves a \(1.66\times\) decoding speedup over Dense.

Our contributions are summarized as follows:
\begin{itemize}[leftmargin=1.7em]
    \item We show that an individual query head can exhibit markedly different retrieval behavior at different relative distances, exposing a limitation of treating the entire head as a single unit for cache allocation.

    \item We introduce Distance-KV, a model- and budget-specific static pattern that determines KV retention jointly over layers, query heads, and relative distances, enabling physical KV cache compression without online scoring.
    
    \item We evaluate Distance-KV across three backbone models and four complementary long-context benchmarks. Structural ablations confirm the importance of the learned query-head--distance structure, while pattern analyses reveal model-specific distance-selective retention.
\end{itemize}

\section{Related Work}
\label{sec:related-work}

\paragraph{Efficient long-context inference.}
Long-context inference has been accelerated at the kernel, memory-management, and execution levels. FlashAttention~\citep{dao2022flashattention} reduces memory traffic in exact attention, while PagedAttention~\citep{10.1145/3600006.3613165} manages KV caches through noncontiguous memory blocks. MInference~\citep{3737916.3739579} and FlexPrefill~\citep{lai2025flexprefill} exploit attention sparsity to reduce the cost of long-context prefilling. Recurrent and segmented execution methods further improve scalability by carrying bounded states across segments~\citep{dai-etal-2019-transformer,shang2026traininginference}.

\paragraph{KV cache compression.}
KV cache compression reduces storage and decoding costs by retaining only a subset of KV states. StreamingLLM~\citep{xiao2024efficient} preserves fixed sink and recent tokens, while H$_2$O~\citep{zhang2023ho}, SnapKV~\citep{3737916.3738638}, and DefensiveKV~\citep{feng2026defensivekv} use attention-derived scores to retain important tokens. PyramidKV~\citep{cai2025pyramidkv} varies the cache budget across layers, whereas KVzip~\citep{NEURIPS2025_f4eaa4b8} derives query-agnostic KV importance through context reconstruction.

\paragraph{Attention-head heterogeneity and retrieval behavior.}
Attention heads exhibit substantial functional heterogeneity. Work on retrieval heads~\citep{wu2025retrieval} identifies a small subset of heads responsible for long-range retrieval, while FastGen~\citep{ge2024model} assigns cache compression policies according to head behavior. DuoAttention~\citep{xiao2025duoattention} separates retrieval heads from streaming heads, and MoA~\citep{DBLP:journals/corr/abs-2406-14909} searches for sparse attention rules specific to individual layers and heads. Despite their different formulations, these methods assign each head an overall role or parameterized sparse rule rather than directly modeling its retrieval utility across relative-distance intervals. Distance-KV instead models this missing structure through a model- and budget-specific static retention pattern over layers, query heads, and relative distances.

\begin{figure}[t]
    \centering
    \includegraphics[width=\linewidth]{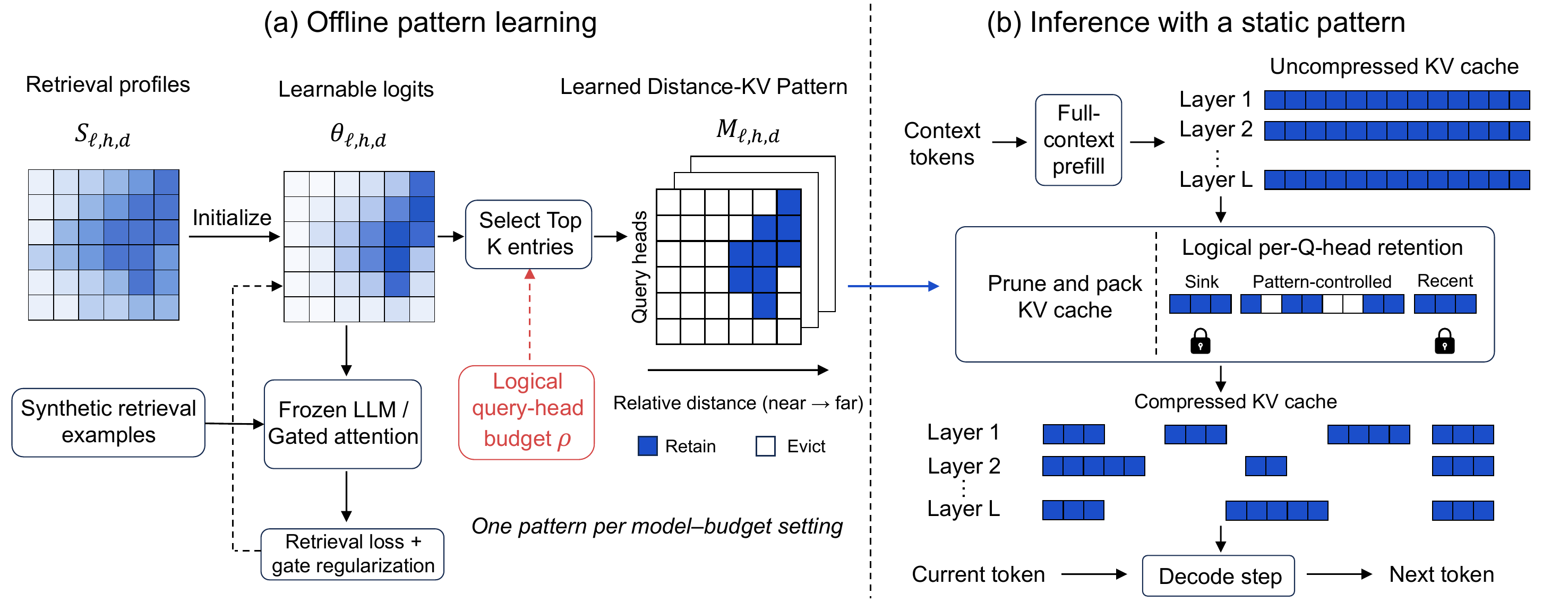}
\caption{Overview of Distance-KV.
(a) Retrieval-guided offline learning produces a model- and budget-specific
binary pattern with one query-head-by-distance matrix per layer while keeping
the language model frozen.
(b) After full-context prefill, the pattern is mapped to the current context
and applied once to prune and pack its KV cache; the same pattern is reused
across inputs.}
\label{fig:method-overview}
\end{figure}

\section{Distance-KV}
\label{sec:method}

Existing retrieval-head methods for KV cache compression typically characterize each attention head using a single importance score or binary label and make cache allocation decisions at the granularity of entire heads. However, Figure~\ref{fig:motivation} shows that a head's retrieval capability can vary with relative distance, so head-level allocation can conflate useful and redundant distance ranges. Motivated by this observation, we propose Distance-KV, which learns a fine-grained static KV retention policy over layers, query heads, and relative distances. Figure~\ref{fig:method-overview} summarizes how the model- and budget-specific pattern is learned offline and reused to prune and pack the KV cache at inference time. We first quantify the distance-dependent retrieval behavior, then present the pattern representation, its budget-constrained learning procedure, and static-pattern inference.

\subsection{Retrieval Capability Varies with Distance}
\label{sec:distance-dependent-retrieval}
To quantify how the retrieval capability of attention heads varies with distance, we probe every query head in a frozen language model using controlled synthetic retrieval examples (see Appendix~\ref{app:retrieval-profile-measurement} for experimental details). In each example, we place the target value at a different relative distance from the query and record the attention distributions of the query heads as the model predicts the tokens of that value. Adapting the retrieval score of \citet{wu2025retrieval}, we define the distance-conditioned retrieval score@5 for query head \(h\) in layer \(\ell\) and distance bin \(d\), denoted by \(S_{\ell,h,d}\), as the proportion of target tokens whose corresponding source positions appear among the five positions receiving the highest attention from that head. 

Figure~\ref{fig:motivation} provides an overall view of how retrieval capability varies with distance across attention heads. In Figure~\ref{fig:motivation}(a), we divide each head's scores by its maximum across distance bins and sort the 1,024 query heads in Llama-3.1-8B-Instruct by the bin in which this maximum occurs. The resulting high-score regions span different distance ranges, revealing distinct distance-dependent retrieval behavior across heads. Since this normalization is used only to highlight within-head variation, Figure~\ref{fig:motivation}(b) further reports the raw retrieval scores of two example heads with contrasting profiles. L2H22 performs better at shorter distances, whereas L21H8 becomes stronger at longer distances. Figure~\ref{fig:motivation}(c) extends this comparison to all query heads by plotting their mean raw retrieval scores over the four farthest distance bins against those over the four nearest bins. Points on both sides of the diagonal show that this contrast extends beyond the two highlighted heads.

The results reveal substantial variation with relative distance across the layer--query-head space. A single head score cannot distinguish the distance ranges in which a head is useful, and the observed profiles are not reducible to a simple near--far dichotomy: some heads peak at intermediate distances or vary nonmonotonically. We therefore formulate KV allocation over individual layer--query-head--distance entries using a Distance-KV Pattern \(M\), formally defined in Section~\ref{sec:pattern-formulation}.

\subsection{Distance-KV Pattern}
\label{sec:pattern-formulation}

To capture the distance-dependent retrieval variation identified in Section~\ref{sec:distance-dependent-retrieval}, we partition the compressible portion of the context into \(D\) relative-distance bins and define a separate retention decision for every layer, query head, and distance bin. For a model with $L$ layers and $H$ query heads per layer, the Distance-KV Pattern is a binary tensor $M\in\{0,1\}^{L\times H\times D}$. Each entry $M_{\ell,h,d}=1$ indicates that the context positions in distance bin $d$ are logically retained for query head $h$ in layer $\ell$, whereas $M_{\ell,h,d}=0$ indicates that they are not. As illustrated in Figure~\ref{fig:method-overview}(a), each layer corresponds to a binary query-head-by-distance matrix, and the collection of these matrices forms the complete Distance-KV Pattern. Here, $M$ defines a logical retention policy over query heads. For GQA models, its mapping to shared physical KV heads is described in Section~\ref{sec:static-inference}.

For each model and retention setting, the Distance-KV Pattern is learned once in relative-distance coordinates and reused across inputs. A distance-agnostic head-level policy is recovered by setting $M_{\ell,h,d}=z_{\ell,h}$ for all $d$. More generally, each query head can retain arbitrary, potentially disjoint distance regions, rather than being restricted to a single sliding window. Sink and recent tokens are always retained, while $M_{\ell,h,d}$ controls the compressible context between them.

\subsection{Budget-Constrained Pattern Learning}
\label{sec:pattern-learning}

As illustrated in Figure~\ref{fig:method-overview}(a), given a model and a target logical query-head retention ratio $\rho$, Distance-KV learns the corresponding retention pattern offline from synthetic retrieval examples. During training, all parameters of the pretrained language model remain frozen, and only the gate logits $\theta_{\ell,h,d}$ associated with the layer--query-head--distance entries are optimized. We define $\mathcal{L}_{\mathrm{ret}}(M)$ as the teacher-forced cross-entropy on the target-value tokens of the synthetic retrieval examples when the frozen model uses pattern $M$. We initialize each gate probability from the corresponding distance-conditioned retrieval score $S_{\ell,h,d}$. Let \(K_{\rho}=\operatorname{round}\!\left(\rho LHD\right)\) denote the number of active entries under the target retention ratio. We formulate the desired binary pattern as
\begin{equation}
\label{eq:budget-objective}
\begin{aligned}
M_{\rho}^{\star}
=
\underset{M}{\operatorname{arg\,min}}
\quad & \mathcal{L}_{\mathrm{ret}}(M) \\
\text{s.t.}\quad
& M\in\{0,1\}^{L\times H\times D},
\quad
\lVert M\rVert_{0}=K_{\rho}.
\end{aligned}
\end{equation}
To optimize this discrete objective, we first use Hard Concrete gates~\citep{louizos2018learning} to learn a continuous relaxation of the retention variables and then refine an exact-budget binary pattern through straight-through Top-K optimization.

In the first stage, we relax the discrete retention decisions using Hard Concrete gates. Each gate logit $\theta_{\ell,h,d}$ parameterizes a logical query-head retention variable $Z_{\ell,h,d}$. During the forward pass, the resulting relaxed gates modulate the attention mass over the context positions associated with each distance bin. Let \(p_{\ell,h,d}=\Pr\!\left(Z_{\ell,h,d}>0\right)\) denote the probability that a pattern entry is active. We optimize
\begin{equation}
\label{eq:hard-concrete-objective}
\mathcal{L}_{\mathrm{HC}}
=
\mathcal{L}_{\mathrm{ret}}(Z)
+
\frac{\lambda}{LHD}
\sum_{\ell,h,d}p_{\ell,h,d}.
\end{equation}
The first term preserves retrieval performance on the synthetic examples, while the second is an expected $L_0$ regularizer that promotes sparsity. This stage learns differentiable retention probabilities without imposing the exact target ratio. Additional details of the initialization and relaxed gates are provided in Appendix~\ref{app:hard-concrete}.

In the second stage, we initialize from the relaxed solution. Let $P(\theta)\in[0,1]^{L\times H\times D}$ collect the nonzero probabilities, with $P_{\ell,h,d}(\theta)=p_{\ell,h,d}$. We construct an exact-budget binary pattern using global Top-K selection:
\begin{equation}
\label{eq:topk-pattern}
M_{\rho}(\theta)
=
\operatorname{TopKMask}\left(P(\theta),K_{\rho}\right),
\end{equation}
where $\operatorname{TopKMask}(\cdot,K_{\rho})$ sets the $K_{\rho}$ largest entries to one and all remaining entries to zero. Because the Top-K operator is nondifferentiable, we use a straight-through estimator~\citep{bengio2013estimating}: the forward pass follows this exact-budget binary pattern, while gradients are propagated to the gate logits through the relaxed Hard Concrete attention path. After optimization, we apply the same global Top-K selection and export the resulting deterministic Distance-KV Pattern. Details of the straight-through estimator and global Top-K selection are provided in Appendix~\ref{app:straight-through-topk}.

\subsection{Static-Pattern Inference}
\label{sec:static-inference}
At inference time, given a context of length $T_c$, we map the relative-distance bins in the learned Distance-KV Pattern to the corresponding context positions, using the end of the context as the reference point. For a context token at position $i\in\{129,\ldots,T_c-1024\}$, its zero-indexed relative-distance bin is
\begin{equation}
\label{eq:distance-bin}
d(i;T_c)
=
\left\lfloor\frac{T_c-i-1024}{128}\right\rfloor.
\end{equation}
The logical retention decision for query head $h$ in layer $\ell$ at position $i$ is then
\begin{equation}
\label{eq:retention-mask}
m_{\ell,h,i}
=
\begin{cases}
1,
& i\in\mathcal{S}(T_c)\cup\mathcal{R}(T_c),\\
M_{\ell,h,d(i;T_c)},
& \text{otherwise}.
\end{cases}
\end{equation}
Here, $\mathcal{S}(T_c)$ and $\mathcal{R}(T_c)$ denote the first 128 sink positions and the most recent 1,024 positions, respectively; both are always retained. The learned pattern controls only the compressible context between these two regions. Because the mapping uses relative rather than absolute positions, only the bins covered by the current context are instantiated, allowing the same pattern to support different context lengths.

As illustrated in Figure~\ref{fig:method-overview}(b), after dense prefill over the context, we apply the retention decisions defined above to perform a one-time static compression of the context KV cache. The KV states marked for eviction are physically removed, while the remaining states are packed in their original token order. The packing operation changes only the storage layout of the retained KV states and preserves their original position indices and RoPE semantics. The query tokens then attend to the compressed context cache, while the KV states of the query tokens and subsequently generated tokens are appended normally and retained in full. Distance-KV neither recomputes nor reapplies the pattern during generation and therefore requires no query-dependent importance scoring or online pattern optimization.

For MHA, the logical retention decision of each query head applies directly to its corresponding physical KV head. For GQA, the decisions of query heads sharing the same KV head are combined by taking their union, so a position is retained whenever any query head in the group selects it. Consequently, $\rho$ specifies the logical query-head retention ratio, whereas cache usage and attention cost are governed by the physical KV retention ratio after this union.

\section{Experiments}
\label{sec:experiments}

\subsection{Experimental Setup}
\label{sec:experimental-setup}

\paragraph{Models and benchmarks.}
We evaluate Distance-KV on four complementary long-context benchmarks using three backbone models: Llama-2-7B-32K-Instruct~\citep{DBLP:journals/corr/abs-2307-09288}, Llama-3.1-8B-Instruct~\citep{DBLP:journals/corr/abs-2407-21783}, and Qwen2.5-7B-Instruct~\citep{DBLP:journals/corr/abs-2412-15115}. Needle-in-a-Haystack (NIAH) varies both context length and needle position to assess position-wise retrieval robustness. LongBench~\citep{bai-etal-2024-longbench} covers diverse downstream tasks, RULER~\citep{hsieh2024ruler} measures robustness across context lengths through controlled tasks, and SCBench~\citep{li2025scbench} evaluates KV cache reuse in shared-context, multi-turn settings. 

\paragraph{Baselines and evaluation protocol.}
We compare Distance-KV with the full-cache Dense baseline and four KV cache compression methods: StreamingLLM~\citep{xiao2024efficient}, DuoAttention~\citep{xiao2025duoattention}, MoA~\citep{DBLP:journals/corr/abs-2406-14909}, and the context-independent variant of KVzip~\citep{NEURIPS2025_f4eaa4b8}, denoted as KVzip-CI. All compression methods determine which states to retain from the initial context KV cache before processing the query, without access to the query content. In SCBench, the compressed shared-context KV cache is reused across subsequent queries. We follow the official prompt construction, generation, and evaluation protocols of LongBench, RULER, and SCBench.

\paragraph{Implementation details.}
Distance-KV uses a logical Q-head retention ratio of \(20\%\) in all main experiments. For GQA models, the resulting physical KV usage is measured after unioning the decisions of query heads that share a KV head, as described in Section~\ref{sec:static-inference}, and is reported in Section~\ref{sec:efficiency} and Appendix~\ref{app:additional-efficiency}.
We learn one model-specific pattern for each backbone and reuse it unchanged across all benchmarks evaluated with that model. 
The packed KV-cache update kernel used during inference is adapted from DefensiveKV~\citep{feng2026defensivekv}. All baselines are evaluated using their publicly released implementations with fixed model-specific configurations across benchmarks. Additional pattern-learning details are provided in Appendix~\ref{app:pattern-learning}, and baseline configurations are provided in Appendix~\ref{app:baseline-configurations}. Our code is available at \href{https://github.com/sxp11/Distance-KV-Pattern}{link}.

\subsection{Long-Context Performance}
\label{sec:main-results}

\begin{figure}[h]
    \centering
    \includegraphics[width=\linewidth]{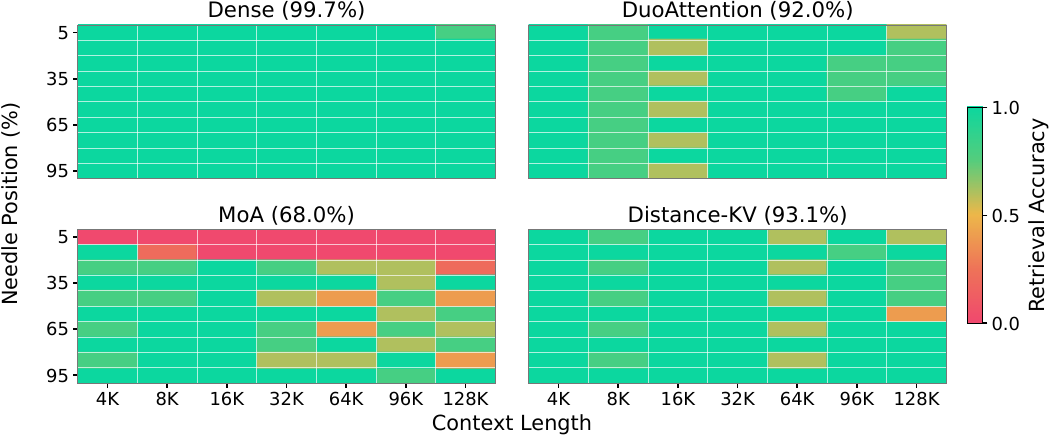}
    \caption{Retrieval accuracy of Llama-3.1-8B-Instruct on multi-needle NIAH across context lengths and needle positions. Values in parentheses report the mean accuracy over the full evaluation grid.}
\label{fig:niah}
\end{figure}

\paragraph{Position-wise retrieval robustness.}
Figure~\ref{fig:niah} shows that Distance-KV preserves high retrieval accuracy over nearly the entire length--position grid, including 128K contexts. MoA exhibits broad low-accuracy regions, whereas DuoAttention remains competitive but shows more localized degradation. A single static pattern therefore supports retrieval across widely varying relative distances.

\paragraph{Downstream long-context performance.}
As shown in Table~\ref{tab:longbench-v1}, Distance-KV achieves the highest average LongBench score among all KV cache compression methods on each of the three backbones. It is also the best-performing compression method on multi-document question answering for all three models, a category that requires evidence integration across documents. The patterns learned from synthetic retrieval therefore transfer to natural long-document tasks and remain effective on both MHA and GQA backbones. Complete task-level results are reported in Appendix~\ref{app:longbench-results}.

\begin{figure}[h]
    \centering
    \includegraphics[width=\linewidth]{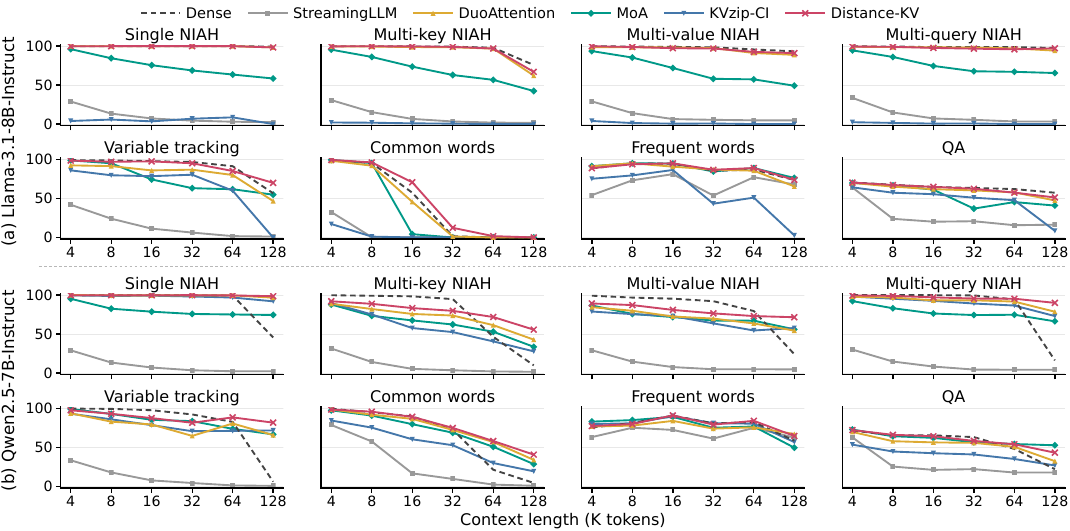}
    \caption{\textbf{RULER task-group scores across context lengths.}
Results are shown for (a) Llama-3.1-8B-Instruct and
(b) Qwen2.5-7B-Instruct from 4K to 128K tokens.
For task groups containing multiple tasks, we report the mean score
within each group.}
\label{fig:ruler}
\end{figure}

\begin{table}[h]
\centering
\caption{LongBench performance across three backbone models. Scores are averaged within each of the six official task categories, and Avg. denotes the overall average across all 21 datasets. Bold indicates the best KV cache compression method in each column.}
\label{tab:longbench-v1}
\vspace{0.3em}
\begingroup
\setlength{\tabcolsep}{2.4pt}
\renewcommand{\arraystretch}{0.92}

\begin{tabularx}{\linewidth}{@{}ll*{7}{Y}}
\toprule[0.09em]
\multirow{2}{*}{\hspace{0.8em}\textbf{Model}}
& \multirow{2}{*}{\hspace{0.8em}\textbf{Method}}
& \multicolumn{7}{c}{\textbf{LongBench}} \\
\cmidrule(lr){3-9}
& & \textbf{S-Doc} & \textbf{M-Doc} & \textbf{Summ.}
& \textbf{Few.} & \textbf{Synth.} & \textbf{Code} & \textbf{Avg.} \\
\midrule[0.075em]

\multirow{6}{*}{\shortstack[c]{Llama-2-7B-\\[-1pt]32K-Instruct}}
& Dense
& 12.32 & 12.59 & 15.18 & 56.90 & 7.19 & 62.64 & 25.47 \\
\cmidrule(lr){2-9}
& StreamingLLM
& 8.42 & 9.12 & 13.51 & 36.13 & 3.11 & 57.71 & 18.74 \\
& DuoAttention
& \textbf{13.82} & 11.57 & 14.74 & 48.24 & 5.61 & 54.48 & 22.82 \\
& MoA
& 11.77 & 10.90 & 13.75 & \textbf{55.86} & \textbf{9.16} & 61.21 & 24.72 \\
& KVzip-CI
& 8.85 & 9.15 & 11.73 & 30.47 & 2.25 & 50.36 & 16.59 \\
\rowcolor{distancekvblue}
\cellcolor{white}
& \textbf{Distance-KV}
& 12.58 & \textbf{12.63} & \textbf{14.76} & 54.34 & 8.70
& \textbf{61.56} & \textbf{25.07} \\
\midrule[0.075em]

\multirow{6}{*}{\shortstack[c]{Llama-3.1-8B-\\[-1pt]Instruct}}
& Dense
& 48.58 & 41.92 & 17.75 & 61.53 & 66.34 & 60.01 & 47.53 \\
\cmidrule(lr){2-9}
& StreamingLLM
& 23.96 & 24.26 & 15.88 & 41.60 & 14.00 & 60.42 & 27.89 \\
& DuoAttention
& 45.77 & 39.48 & 17.51 & \textbf{60.53} & \textbf{67.25}
& \textbf{65.25} & 46.92 \\
& MoA
& 43.50 & 35.51 & 17.21 & 59.64 & 57.39 & 62.76 & 43.86 \\
& KVzip-CI
& 43.49 & 31.70 & 14.60 & 55.03 & 53.98 & 19.66 & 37.17 \\
\rowcolor{distancekvblue}
\cellcolor{white}
& \textbf{Distance-KV}
& \textbf{48.39} & \textbf{41.08} & \textbf{17.76} & 60.44
& 66.72 & 61.20 & \textbf{47.30} \\
\midrule[0.075em]

\multirow{6}{*}{\shortstack[c]{Qwen2.5-7B-\\[-1pt]Instruct}}
& Dense
& 45.97 & 39.09 & 16.42 & 56.86 & 62.17 & 60.36 & 44.79 \\
\cmidrule(lr){2-9}
& StreamingLLM
& 23.07 & 22.70 & 15.48 & 44.64 & 13.00 & 55.24 & 27.29 \\
& DuoAttention
& 43.36 & 36.11 & \textbf{16.85} & \textbf{56.20} & 58.83
& 61.19 & 43.28 \\
& MoA
& 40.41 & 34.27 & 16.48 & 56.06 & 56.50 & \textbf{62.25} & 42.04 \\
& KVzip-CI
& 33.44 & 29.11 & 14.36 & 50.16 & 29.72 & 44.40 & 32.68 \\
\rowcolor{distancekvblue}
\cellcolor{white}
& \textbf{Distance-KV}
& \textbf{43.91} & \textbf{37.90} & 16.46 & 55.68
& \textbf{62.17} & 60.85 & \textbf{44.00} \\
\bottomrule[0.09em]

\end{tabularx}
\endgroup
\end{table}

\paragraph{Robustness across context lengths.}
Figure~\ref{fig:ruler} shows that the separation among methods is clearest at long context lengths. Averaged over the eight task groups at 128K, Distance-KV outperforms the strongest compression baseline by 5.7 points on Llama-3.1-8B-Instruct and 9.3 points on Qwen2.5-7B-Instruct. The gains extend beyond the four NIAH variants to variable tracking, aggregation, and question answering, indicating that the benefit of distance-conditioned retention is not limited to exact retrieval. Complete task-level results at 128K are reported in Appendix~\ref{app:ruler-results}.

\begin{table}[h]
\centering
\caption{SCBench multi-turn performance on two backbone models across
12 tasks grouped into four capability categories.
All scores are reported on a 0--100 scale, with higher values indicating
better performance. Avg.\ denotes the unweighted mean of the four category
averages, and bold indicates the best KV cache compression method in each column.}
\label{tab:scbench}
\vspace{0.3em}
\begingroup
\setlength{\tabcolsep}{1.0pt}
\renewcommand{\arraystretch}{1.06}

\begin{tabularx}{\linewidth}{@{}L*{13}{c}}
\toprule[0.09em]
\multirow{2}{*}{\textbf{Method}}
& \multicolumn{3}{c}{\textbf{String retrieval}}
& \multicolumn{4}{c}{\textbf{Semantic retrieval}}
& \multicolumn{3}{c}{\textbf{Global}}
& \multicolumn{2}{c}{\textbf{Multi-task}}
& \multirow{2}{*}{\textbf{Avg.}} \\
\cmidrule(lr){2-4}
\cmidrule(lr){5-8}
\cmidrule(lr){9-11}
\cmidrule(lr){12-13}
& \textbf{KV} & \textbf{P--S} & \textbf{VT}
& \textbf{Repo} & \textbf{En.QA} & \textbf{Zh.QA} & \textbf{Choice}
& \textbf{ICL} & \textbf{Summ.} & \textbf{Math}
& \textbf{S+N} & \textbf{R+KV} & \\
\midrule[0.075em]

\multicolumn{14}{@{}l}{\textbf{Llama-3.1-8B-Instruct}} \\[3pt]

{\small Dense}
& 80.4 & 41.6 & 47.6 & 50.5 & 29.6 & 20.5 & 69.4
& 43.3 & 40.4 & 21.8 & 61.2 & 68.9 & 49.8 \\
\cmidrule(lr){1-14}

{\small StreamingLLM}
& 0.2 & 0.2 & 0.0 & 0.2 & 5.7 & 6.4 & 26.2
& \textbf{72.2} & 25.3 & 5.4 & 30.3 & 0.2 & 14.8 \\

{\small DuoAttention}
& \textbf{58.4} & 19.6 & \textbf{43.0} & \textbf{42.1}
& 25.9 & 16.9 & 65.1 & 58.9 & 35.8 & 9.4 & 49.7
& \textbf{45.5} & 40.0 \\

{\small MoA}
& 26.2 & 13.0 & 11.4 & 20.5 & 16.2 & 15.6 & 66.8
& 61.9 & 31.4 & 9.8 & 42.5 & 14.8 & 27.4 \\

{\small KVzip-CI}
& 0.0 & 0.0 & 1.9 & 7.7 & 5.1 & 1.8 & 18.8
& 57.8 & 19.3 & 0.2 & 20.4 & 4.6 & 11.8 \\

\rowcolor{distancekvblue}
{\small\textbf{Distance-KV}}
& 56.8 & \textbf{20.8} & 28.8 & 38.6 & \textbf{27.1}
& \textbf{20.2} & \textbf{70.3} & 59.6 & \textbf{36.6}
& \textbf{20.4} & \textbf{58.2} & 43.2 & \textbf{41.0} \\
\midrule[0.075em]

\multicolumn{14}{@{}l}{\textbf{Qwen2.5-7B-Instruct}} \\[3pt]

{\small Dense}
& 12.4 & 26.8 & 36.8 & 44.5 & 23.1 & 4.1 & 63.3
& 57.8 & 33.7 & 23.2 & 65.6 & 44.1 & 38.0 \\
\cmidrule(lr){1-14}

{\small StreamingLLM}
& 0.2 & 0.0 & 0.0 & 0.2 & 6.3 & 1.8 & 24.5
& 70.7 & 27.5 & 5.2 & 37.7 & 0.0 & 15.4 \\

{\small DuoAttention}
& 2.8 & 4.2 & 18.5 & 24.5 & 19.9 & 3.1 & 55.0
& \textbf{71.9} & 29.4 & 16.2 & 63.2 & 18.9 & 28.6 \\

{\small MoA}
& 6.6 & 4.8 & 8.9 & 25.7 & 18.8 & 2.8 & 55.9
& 71.5 & 29.3 & 10.6 & 51.9 & 14.1 & 25.7 \\

{\small KVzip-CI}
& 1.2 & 0.0 & 24.6 & 12.3 & 16.5 & \textbf{3.3}
& 52.0 & 58.5 & 25.5 & 5.8 & 45.6 & 9.1 & 21.7 \\

\rowcolor{distancekvblue}
{\small\textbf{Distance-KV}}
& \textbf{15.0} & \textbf{13.2} & \textbf{30.0}
& \textbf{37.5} & \textbf{24.5} & 2.6 & \textbf{62.0}
& 64.8 & \textbf{32.3} & \textbf{22.6}
& \textbf{64.7} & \textbf{27.3} & \textbf{34.2} \\
\bottomrule[0.09em]

\end{tabularx}
\endgroup

\begin{minipage}{\linewidth}
\footnotesize
\raggedright
\textit{Task abbreviations:}
KV, key--value retrieval;
P--S, prefix--suffix matching;
VT, variable tracking;
Repo, repository QA;
S+N, summarization with NIAH;
R+KV, repository QA with KV retrieval.
\end{minipage}
\end{table}

\paragraph{Shared-context multi-turn performance.}
Distance-KV achieves the highest average among compression methods on both backbones (Table~\ref{tab:scbench}), ranking first on 7 of 12 Llama-3.1 tasks and 10 of 12 Qwen2.5 tasks, with wins in every capability category. A single fixed pattern therefore remains effective across successive queries without query-dependent rescoring.

\subsection{Efficiency}
\label{sec:efficiency}

\begin{figure}[h]
    \centering
    \includegraphics[width=\linewidth]{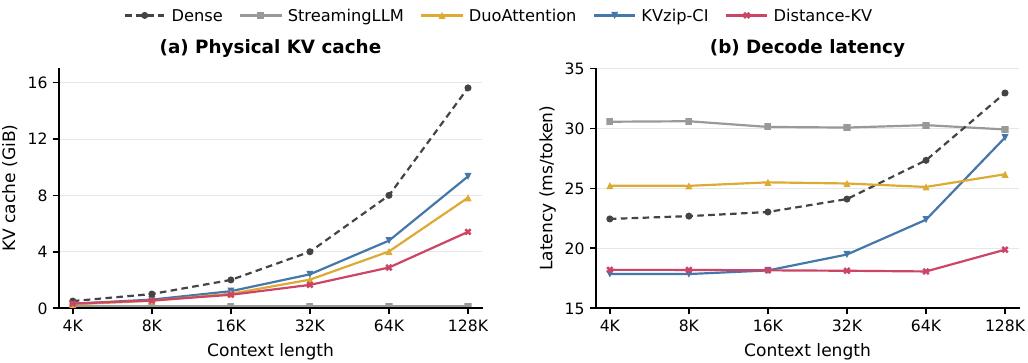}
    \caption{\textbf{Inference efficiency across context lengths.}
(a) Physical KV cache size and (b) per-token decoding latency for
Llama-3.1-8B-Instruct on one NVIDIA H100 GPU with batch size 1.}
\label{fig:efficiency}
\end{figure}

\paragraph{Inference efficiency.}
Figure~\ref{fig:efficiency} compares the physical KV cache size and per-token decoding latency as the context length increases. We omit MoA because its optimized kernel operates on an MHA-converted checkpoint rather than the native GQA backbone. At 128K, Distance-KV uses 5.41 GiB for the context KV cache, 65.4\% less than Dense and also less than DuoAttention and KVzip-CI. Its decoding latency stays close to 18 ms/token from 4K through 64K and rises to only 19.88 ms/token at 128K, the lowest at this length among the compared methods and a $1.66\times$ speedup over Dense. The compacted cache therefore reduces both storage and decoding latency as the context grows. Appendix~\ref{app:additional-efficiency} reports results for Qwen2.5-7B-Instruct and peak GPU memory.

\subsection{Ablation Studies}
\label{sec:ablations}

\begin{table}[h]
\centering
\caption{Ablation results on RULER at 128K for
Llama-3.1-8B-Instruct.
Single, M-Key, and QA average three, three, and two constituent tasks,
respectively; Avg.\ is the mean over all 13 tasks.
Bold indicates the best result in each column. All variants are evaluated at the same training step.}
\label{tab:ruler-ablation}
\vspace{0.3em}
{
\setlength{\tabcolsep}{4.2pt}
\renewcommand{\arraystretch}{1.08}
\begin{tabular}{@{}lccccccccc@{}}
\toprule
\multirow{2}{*}{Variant}
& \multicolumn{4}{c}{NIAH}
& \multirow{2}{*}{VT}
& \multirow{2}{*}{CWE}
& \multirow{2}{*}{FWE}
& \multirow{2}{*}{QA}
& \multirow{2}{*}{Avg.} \\
\cmidrule(lr){2-5}
& Single & M-Key & M-Value & M-Query
& & & & & \\
\midrule
Head-only
& 73.8 & 25.5 & 84.8 & 87.2
& 2.6 & 0.0 & 59.9 & 22.9 & 44.5 \\

Distance-only
& 10.0 & 4.4 & 16.3 & 13.8
& 3.6 & 0.1 & 55.8 & 16.0 & 12.7 \\

Distance-shuffled
& 84.7 & 36.9 & 90.1 & 92.3
& 33.6 & 0.1 & 64.5 & 32.8 & 54.7 \\

\midrule
\textbf{Distance-KV}
& \textbf{95.9} & \textbf{45.5} & \textbf{93.2} & \textbf{95.5}
& \textbf{61.4} & \textbf{0.2} & \textbf{70.3} & \textbf{41.8}
& \textbf{63.7} \\
\bottomrule
\end{tabular}
}
\end{table}

We conduct structural ablations on RULER at 128K with matched physical KV retention across variants. As shown in Table~\ref{tab:ruler-ablation}, Distance-KV performs best across all eight task groups, averaging 63.7. Head-only applies one retention decision to all distance bins within each head and drops by 19.2 points, while Distance-only shares distance-conditioned decisions across query heads within each layer and drops by 51.0 points. Neither whole-head retention nor head-independent distance allocation is therefore sufficient. Distance-shuffled preserves per-head retention counts and physical KV cost but permutes distance-bin assignments, causing a 9.0-point drop. This isolates the learned layer--query-head--distance correspondence as key to Distance-KV's performance. Appendix~\ref{app:additional-ablation} provides complete task-level results, including the fully shuffled control.

\subsection{Analysis of Learned Patterns}
\label{sec:pattern-analysis}

\begin{figure}[h]
    \centering
    \includegraphics[width=\linewidth]{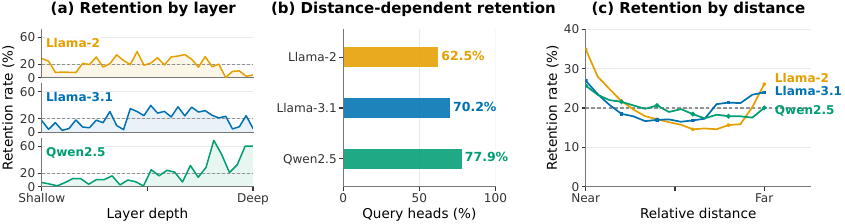}
    \caption{\textbf{Learned Distance-KV patterns across three backbone models.}
All patterns use a 20\% logical query-head retention ratio.
(a) Mean retention rate by layer.
(b) Fraction of query heads that retain only a subset of relative-distance bins.
(c) Mean retention rate by relative distance.
Gray dashed lines in (a) and (c) indicate the 20\% retention ratio.}
\label{fig:pattern-analysis}
\end{figure}

Despite sharing the same overall retention ratio of 20\%, the three models learn distinct, nonuniform layer-wise allocations (Figure~\ref{fig:pattern-analysis}(a)). Specifically, across all layers, 62.5\%, 70.2\%, and 77.9\% of query heads in Llama-2, Llama-3.1, and Qwen2.5, respectively, retain only a subset of distance bins, showing that most heads use distance-dependent retention rather than a single decision across all distances (Figure~\ref{fig:pattern-analysis}(b)). Retention is also distance-dependent: all three models favor nearby positions; Llama-2 and Llama-3.1 increase retention again at long distances, whereas Qwen2.5 has a flatter profile (Figure~\ref{fig:pattern-analysis}(c)). Together with Figure~\ref{fig:motivation}, these patterns establish relative distance as a distinct dimension of retrieval behavior and KV allocation.

\section{Conclusion}
\label{sec:conclusion}
We show that retrieval capability varies substantially with relative distance, even within the same query head. Building on this observation, we introduce Distance-KV, which uses a reusable static retention pattern defined jointly over layers, query heads, and relative distances. Experiments across multiple backbone models show that Distance-KV consistently achieves the best overall performance among the evaluated KV cache compression methods while reducing physical KV cache size and decoding latency. Ablations and learned-pattern analyses further establish the importance of the correspondence between query heads and distance ranges. More broadly, our findings identify relative distance as an important structural dimension for understanding long-context retrieval and motivate further study.

\subsection*{AI use statement}

Generative AI tools were used to assist with implementing portions of the
research code, drafting parts of the manuscript, and polishing the language
for clarity, concision, and readability. All AI-assisted code was reviewed
and tested by the authors, and all AI-assisted text was reviewed and revised
by the authors. The authors take full responsibility for the final content
of this work.

\subsection*{Reproducibility statement}

To facilitate reproducibility, Section~\ref{sec:method} describes Distance-KV,
Section~\ref{sec:experimental-setup} specifies the experimental protocols and
implementation settings, and Appendices~\ref{app:pattern-learning}--%
\ref{app:additional-ablation} provide additional implementation details and
experimental results. The implementation is publicly available via the link in
Section~\ref{sec:experimental-setup}.

\bibliography{iclr2027_conference}
\bibliographystyle{iclr2027_conference}

\appendix
\numberwithin{equation}{section}
\section{Pattern-Learning Details}
\label{app:pattern-learning}
For pattern learning, we reserve 256 tokens of each model's configured sequence-length budget for the query and answer, leaving 32,512 context tokens for Llama-2 and 130,816 for Llama-3.1 and Qwen2.5. We reserve the first 128 context tokens as sink tokens and the most recent 1,024 context tokens as recent tokens, and partition the remaining compressible context into 128-token relative-distance bins. This yields \(D=245\) for Llama-2 and \(D=1{,}013\) for Llama-3.1 and Qwen2.5.

\subsection{Retrieval-Profile Measurement}
\label{app:retrieval-profile-measurement}

Figure~\ref{fig:motivation} uses retrieval scores collected for retrieval-guided pattern initialization with Llama-3.1-8B-Instruct. We insert four key--value records at controlled relative distances into a context of unrelated filler text. Each context uses either access codes consisting of three three-digit groups or verification phrases consisting of three words. Each record is queried separately. Representative records from the two value-type conditions are:
\begin{quote}
\small
\noindent\emph{Numerical:} The access code associated with record key ``harbor'' is ``470 668 610''.\\
\emph{Word:} The verification phrase associated with record key ``harbor'' is ``antique amber armchair''.
\end{quote}
We use two single-key query templates:
\begin{quote}
\small
\noindent\emph{Q1:} The document contains record-key-to-\texttt{\{record\_kind\}} entries. Retrieve the \texttt{\{item\}} associated with record key ``\texttt{\{key\}}''.\\
\emph{Q3:} According to the \texttt{\{record\_kind\}} records in the document, what \texttt{\{item\}} corresponds to record key ``\texttt{\{key\}}''?
\end{quote}
Here, \texttt{\{record\_kind\}} is \emph{access-code} or \emph{verification-phrase}, and \texttt{\{item\}} is the corresponding value name. Both templates additionally specify the value format and instruct the model to output only that value.

The probe uses four distance-coverage rounds. For each distance bin, we average repeated examples within each value-type--template--round condition, then average the condition means equally. For visualization, we aggregate the 1,013 original 128-token distance bins into 16 coarser bins. The near and far scores in Figure~\ref{fig:motivation}(c) average the first and last four of these bins, respectively.

\subsection{Retrieval-Guided Initialization and Relaxed Gates}
\label{app:hard-concrete}

Distance-KV associates each layer--query-head--distance entry with a trainable gate logit $\theta_{\ell,h,d}$. Let $\bar{S}$ denote the mean of the distance-conditioned retrieval scores over all entries. We map each score to an initial nonzero probability as
\begin{equation}
\label{eq:init-retention-probability}
\pi_{\ell,h,d}^{(0)}
=
\operatorname{clip}_{[p_{\min},p_{\max}]}
\left[
\sigma\left(
\operatorname{logit}(\mu)
+
a\left(S_{\ell,h,d}-\bar{S}\right)
\right)
\right],
\end{equation}
where $\mu$ specifies the central retention probability, $a$ controls the dependence on the retrieval score, and $p_{\min}$ and $p_{\max}$ bound the resulting probabilities.

We parameterize each relaxed gate $Z_{\ell,h,d}$ using the Hard Concrete distribution~\citep{louizos2018learning}, with temperature $\beta$ and stretch bounds $\gamma$ and $\zeta$. Its nonzero probability is
\begin{equation}
\label{eq:hard-concrete-probability}
p_{\ell,h,d}
=
\Pr\left(Z_{\ell,h,d}>0\right)
=
\sigma\left(
\theta_{\ell,h,d}
-
\beta\log\frac{-\gamma}{\zeta}
\right).
\end{equation}
We initialize the gate logits by inverting this relation:
\begin{equation}
\label{eq:init-gate-logit}
\theta_{\ell,h,d}^{(0)}
=
\operatorname{logit}\left(\pi_{\ell,h,d}^{(0)}\right)
+
\beta\log\frac{-\gamma}{\zeta}.
\end{equation}

For GQA models, the relaxed gates remain defined separately for query heads that share the same KV head and independently modulate their corresponding attention distributions during pattern learning. The expected \(L_0\) regularizer is likewise defined over these logical query-head gates. The binary retention decisions are combined by union only when the learned logical pattern is mapped to shared physical KV heads at inference, as described in Section~\ref{sec:static-inference}.

\subsection{Exact-Budget Refinement}
\label{app:straight-through-topk}

Let $P(\theta)\in[0,1]^{L\times H\times D}$ denote the tensor of nonzero probabilities, with $P_{\ell,h,d}(\theta)=p_{\ell,h,d}$. For a target logical retention ratio $\rho$, we construct an exact-budget binary pattern as
\begin{equation}
\label{eq:exact-budget-pattern}
K_{\rho}
=
\operatorname{round}\left(\rho LHD\right),
\qquad
M_{\rho}(\theta)
=
\operatorname{TopKMask}\left(P(\theta),K_{\rho}\right),
\end{equation}
where the Top-K selection is performed globally over all layer--query-head--distance entries.

Let $A^{\mathrm{rel}}$ and $A^{\mathrm{bin}}$ denote the normalized attention probabilities produced by the relaxed gates and the binary pattern, respectively. We define the straight-through attention probabilities as
\begin{equation}
\label{eq:straight-through-attention}
A^{\mathrm{ST}}
=
A^{\mathrm{rel}}
+
\operatorname{stopgrad}
\left(
A^{\mathrm{bin}}-A^{\mathrm{rel}}
\right).
\end{equation}
Consequently, the forward pass follows the binary pattern, while the gradients with respect to the gate logits follow the relaxed attention path:
\begin{equation}
\label{eq:straight-through-gradient}
A^{\mathrm{ST}}
\overset{\mathrm{forward}}{=}
A^{\mathrm{bin}},
\qquad
\frac{\partial A^{\mathrm{ST}}}
{\partial\theta_{\ell,h,d}}
=
\frac{\partial A^{\mathrm{rel}}}
{\partial\theta_{\ell,h,d}}.
\end{equation}
The straight-through estimator is applied to the normalized attention probabilities rather than directly to the gate tensor. The refinement objective is
\begin{equation}
\label{eq:straight-through-objective}
\mathcal{L}_{\mathrm{ST}}
=
\mathcal{L}_{\mathrm{ret}}
\left(A^{\mathrm{ST}}\right)
+
\frac{\lambda}{LHD}
\sum_{\ell,h,d}p_{\ell,h,d}.
\end{equation}
After optimization, we apply the same global Top-K selection to the learned nonzero probabilities and export the resulting binary pattern for inference.

\paragraph{Hyperparameters.}
Table~\ref{tab:pattern-learning-hparams} lists the shared hyperparameters for pattern learning.

\begin{table}[t]
\centering
\caption{Shared hyperparameters for offline pattern learning.}
\label{tab:pattern-learning-hparams}
\begin{tabular}{lc}
\toprule
Hyperparameter & Value \\
\midrule
Hard Concrete temperature $\beta$ & $2/3$ \\
Stretch bounds $(\gamma,\zeta)$ & $(-0.1, 1.1)$ \\
Initialization center $\mu$ & $0.7$ \\
Initialization slope $a$ & $6$ \\
Probability bounds $[p_{\min},p_{\max}]$ & $[0.001, 0.999]$ \\
Optimizer & AdamW \\
Learning rate & $0.01$ \\
AdamW weight decay & $0$ \\
Gradient clipping norm & $1.0$ \\
Final $L_0$ coefficient $\lambda$ & $0.5$ \\
$L_0$ ramp steps & $4$ \\
\bottomrule
\end{tabular}
\end{table}

\section{Baseline Configurations}
\label{app:baseline-configurations}

All baselines use their publicly released implementations with fixed model-specific configurations across benchmarks. When an author-provided retention pattern or sparse attention configuration is available for a backbone, we use it directly; otherwise, we derive a model-specific configuration using the official implementation and search procedure. Dense retains the full KV cache. StreamingLLM retains 128 sink tokens and 1,024 recent tokens. DuoAttention retains 64 sink tokens and 256 recent tokens, with a sparsity ratio of 0.75 for Llama-2-7B-32K-Instruct and 0.50 for the other two backbones. MoA uses a fixed model-specific sparse attention configuration with a density of 0.50. For KVzip, we use its context-independent variant, denoted as KVzip-CI, with a head-level pruning ratio of 0.60.

\section{Additional Benchmark Results}
\label{app:additional-results}

\subsection{Task-Level LongBench Results}
\label{app:longbench-results}

Tables~\ref{tab:longbench-full-llama2}--\ref{tab:longbench-full-qwen25}
report the complete results on all 21 LongBench tasks. These task-level
results are aggregated into the six category scores reported in
Table~\ref{tab:longbench-v1}.

\begin{table}[h]
\centering
\caption{Task-level LongBench results on Llama-2-7B-32K-Instruct. Avg. is the unweighted mean over all 21 tasks. Bold indicates the best KV cache compression method in each row.}
\label{tab:longbench-full-llama2}
\setlength{\tabcolsep}{2.8pt}
\renewcommand{\arraystretch}{0.94}
\begin{tabular*}{\linewidth}{@{\extracolsep{\fill}}lcccccc@{}}
\toprule
\multirow{2}{*}{\textbf{Task}}
& \multirow{2}{*}{\textbf{Dense}}
& \textbf{Streaming}
& \textbf{Duo}
& \multirow{2}{*}{\textbf{MoA}}
& \multirow{2}{*}{\textbf{KVzip-CI}}
& \textbf{Distance} \\
& & \textbf{LLM} & \textbf{Attention} & & & \textbf{KV} \\
\midrule

\multicolumn{7}{l}{\textit{Single-document QA}} \\
\quad NarrativeQA
& 4.89 & 4.14 & 5.00 & 4.33 & \textbf{6.79} & 3.64 \\
\quad Qasper
& 12.30 & 7.22 & \textbf{15.04} & 12.32 & 6.91 & 11.59 \\
\quad MultiFieldQA-en
& 20.97 & 15.93 & \textbf{23.07} & 20.83 & 17.00 & 21.93 \\
\quad MultiFieldQA-zh
& 11.14 & 6.38 & 12.19 & 9.60 & 4.69 & \textbf{13.18} \\

\addlinespace[2pt]
\multicolumn{7}{l}{\textit{Multi-document QA}} \\
\quad HotpotQA
& 13.58 & 8.21 & 13.01 & \textbf{13.30} & 10.54 & 12.74 \\
\quad 2WikiMQA
& 11.40 & 9.09 & 10.15 & \textbf{12.20} & 8.05 & 11.93 \\
\quad MuSiQue
& 7.08 & 3.91 & 5.67 & 6.84 & 5.25 & \textbf{7.50} \\
\quad DuReader
& 18.29 & 15.27 & 17.47 & 11.27 & 12.76 & \textbf{18.36} \\

\addlinespace[2pt]
\multicolumn{7}{l}{\textit{Summarization}} \\
\quad GovReport
& 19.92 & 15.98 & 19.27 & 19.02 & 17.29 & \textbf{19.42} \\
\quad QMSum
& 11.20 & \textbf{11.45} & 11.01 & 10.05 & 7.97 & 10.17 \\
\quad MultiNews
& 16.04 & 15.08 & 15.56 & 15.35 & 11.49 & \textbf{16.08} \\
\quad VCSUM
& 13.57 & 11.53 & 13.12 & 10.60 & 10.18 & \textbf{13.38} \\

\addlinespace[2pt]
\multicolumn{7}{l}{\textit{Few-shot learning}} \\
\quad TREC
& 72.50 & 46.00 & \textbf{72.50} & 72.00 & 41.00 & 70.50 \\
\quad TriviaQA
& 85.12 & 50.91 & 76.63 & 84.23 & 49.27 & \textbf{84.91} \\
\quad SAMSum
& 35.97 & 34.13 & 32.32 & \textbf{36.21} & 22.20 & 35.96 \\
\quad LSHT
& 34.00 & 13.50 & 11.50 & \textbf{31.00} & 9.42 & 26.00 \\

\addlinespace[2pt]
\multicolumn{7}{l}{\textit{Synthetic tasks}} \\
\quad PassageCount
& 0.67 & 0.17 & \textbf{0.67} & \textbf{0.67} & 0.10 & 0.50 \\
\quad PassageRetrieval-en
& 13.62 & 5.04 & 12.42 & \textbf{22.20} & 4.25 & 16.12 \\
\quad PassageRetrieval-zh
& 7.29 & 4.12 & 3.75 & 4.62 & 2.40 & \textbf{9.47} \\

\addlinespace[2pt]
\multicolumn{7}{l}{\textit{Code completion}} \\
\quad LCC
& 63.62 & 59.28 & \textbf{65.44} & 61.07 & 42.87 & 62.06 \\
\quad RepoBench-P
& 61.67 & 56.14 & 43.52 & \textbf{61.35} & 57.86 & 61.07 \\

\midrule
\textbf{Avg.}
& 25.47 & 18.74 & 22.82 & 24.72 & 16.59 & \textbf{25.07} \\
\bottomrule
\end{tabular*}
\end{table}

\begin{table}[h]
\centering
\caption{Task-level LongBench results on Llama-3.1-8B-Instruct. Avg. is the unweighted mean over all 21 tasks. Bold indicates the best KV cache compression method in each row.}
\label{tab:longbench-full-llama31}
\setlength{\tabcolsep}{2.8pt}
\renewcommand{\arraystretch}{0.94}
\begin{tabular*}{\linewidth}{@{\extracolsep{\fill}}lcccccc@{}}
\toprule
\multirow{2}{*}{\textbf{Task}}
& \multirow{2}{*}{\textbf{Dense}}
& \textbf{Streaming}
& \textbf{Duo}
& \multirow{2}{*}{\textbf{MoA}}
& \multirow{2}{*}{\textbf{KVzip-CI}}
& \textbf{Distance} \\
& & \textbf{LLM} & \textbf{Attention} & & & \textbf{KV} \\
\midrule

\multicolumn{7}{l}{\textit{Single-document QA}} \\
\quad NarrativeQA
& 29.74 & 17.81 & 28.88 & 28.38 & 28.87 & \textbf{31.40} \\
\quad Qasper
& 45.01 & 21.15 & 41.08 & \textbf{44.50} & 42.62 & 44.09 \\
\quad MultiFieldQA-en
& 56.18 & 30.05 & 53.64 & 44.92 & 48.22 & \textbf{55.29} \\
\quad MultiFieldQA-zh
& 63.39 & 26.83 & 59.47 & 56.21 & 54.27 & \textbf{62.77} \\

\addlinespace[2pt]
\multicolumn{7}{l}{\textit{Multi-document QA}} \\
\quad HotpotQA
& 58.01 & 34.62 & 57.12 & 54.27 & 40.38 & \textbf{58.67} \\
\quad 2WikiMQA
& 48.75 & 26.06 & 44.16 & 45.09 & 38.30 & \textbf{47.35} \\
\quad MuSiQue
& 32.93 & 11.57 & 28.20 & 24.21 & 22.91 & \textbf{30.17} \\
\quad DuReader
& 28.00 & 24.78 & \textbf{28.43} & 18.45 & 25.22 & 28.12 \\

\addlinespace[2pt]
\multicolumn{7}{l}{\textit{Summarization}} \\
\quad GovReport
& 19.85 & 16.32 & 19.66 & 18.97 & \textbf{20.16} & 19.89 \\
\quad QMSum
& 19.26 & 15.51 & \textbf{19.30} & 18.26 & 7.43 & 19.06 \\
\quad MultiNews
& 15.46 & 14.92 & 14.95 & 15.24 & 14.85 & \textbf{15.54} \\
\quad VCSUM
& 16.41 & \textbf{16.75} & 16.14 & 16.38 & 15.95 & 16.55 \\

\addlinespace[2pt]
\multicolumn{7}{l}{\textit{Few-shot learning}} \\
\quad TREC
& 73.00 & 48.50 & \textbf{72.50} & \textbf{72.50} & 61.00 & 72.00 \\
\quad TriviaQA
& 92.28 & 68.23 & 89.92 & 89.11 & \textbf{91.33} & 91.08 \\
\quad SAMSum
& 35.36 & 31.16 & 34.71 & 33.45 & 23.28 & \textbf{35.16} \\
\quad LSHT
& 45.50 & 18.50 & \textbf{45.00} & 43.50 & 44.50 & 43.50 \\

\addlinespace[2pt]
\multicolumn{7}{l}{\textit{Synthetic tasks}} \\
\quad PassageCount
& 9.08 & 7.50 & 4.50 & 7.67 & 4.18 & \textbf{8.25} \\
\quad PassageRetrieval-en
& 99.50 & 15.00 & \textbf{99.50} & 68.00 & 97.00 & \textbf{99.50} \\
\quad PassageRetrieval-zh
& 90.45 & 19.50 & \textbf{97.75} & 96.50 & 60.77 & 92.42 \\

\addlinespace[2pt]
\multicolumn{7}{l}{\textit{Code completion}} \\
\quad LCC
& 65.17 & 64.68 & \textbf{69.15} & 65.62 & 9.99 & 65.19 \\
\quad RepoBench-P
& 54.84 & 56.16 & \textbf{61.35} & 59.90 & 29.33 & 57.21 \\

\midrule
\textbf{Avg.}
& 47.53 & 27.89 & 46.92 & 43.86 & 37.17 & \textbf{47.30} \\
\bottomrule
\end{tabular*}
\end{table}

\begin{table}[h]
\centering
\caption{Task-level LongBench results on Qwen2.5-7B-Instruct. Avg. is the unweighted mean over all 21 tasks. Bold indicates the best KV cache compression method in each row.}
\label{tab:longbench-full-qwen25}
\setlength{\tabcolsep}{2.8pt}
\renewcommand{\arraystretch}{0.94}
\begin{tabular*}{\linewidth}{@{\extracolsep{\fill}}lcccccc@{}}
\toprule
\multirow{2}{*}{\textbf{Task}}
& \multirow{2}{*}{\textbf{Dense}}
& \textbf{Streaming}
& \textbf{Duo}
& \multirow{2}{*}{\textbf{MoA}}
& \multirow{2}{*}{\textbf{KVzip-CI}}
& \textbf{Distance} \\
& & \textbf{LLM} & \textbf{Attention} & & & \textbf{KV} \\
\midrule

\multicolumn{7}{l}{\textit{Single-document QA}} \\
\quad NarrativeQA
& 26.85 & 11.29 & \textbf{22.92} & 22.83 & 15.45 & 22.00 \\
\quad Qasper
& 43.30 & 18.34 & 40.38 & \textbf{42.87} & 30.47 & 41.27 \\
\quad MultiFieldQA-en
& 52.09 & 32.85 & 49.53 & 40.34 & 36.96 & \textbf{50.82} \\
\quad MultiFieldQA-zh
& 61.62 & 29.80 & 60.62 & 55.62 & 50.86 & \textbf{61.54} \\

\addlinespace[2pt]
\multicolumn{7}{l}{\textit{Multi-document QA}} \\
\quad HotpotQA
& 57.25 & 32.60 & 52.70 & 52.13 & 41.37 & \textbf{56.23} \\
\quad 2WikiMQA
& 46.47 & 27.35 & 38.59 & 39.75 & 29.20 & \textbf{43.46} \\
\quad MuSiQue
& 30.59 & 10.03 & \textbf{29.31} & 28.83 & 21.77 & 28.85 \\
\quad DuReader
& 22.06 & 20.82 & 23.83 & 16.37 & \textbf{24.12} & 23.06 \\

\addlinespace[2pt]
\multicolumn{7}{l}{\textit{Summarization}} \\
\quad GovReport
& 18.26 & 16.24 & 18.12 & \textbf{18.17} & 11.36 & \textbf{18.17} \\
\quad QMSum
& 18.10 & 14.47 & 17.94 & \textbf{17.98} & 16.06 & 17.88 \\
\quad MultiNews
& 14.73 & 14.59 & 14.68 & \textbf{14.92} & 14.13 & 14.59 \\
\quad VCSUM
& 14.60 & 16.63 & \textbf{16.66} & 14.87 & 15.90 & 15.21 \\

\addlinespace[2pt]
\multicolumn{7}{l}{\textit{Few-shot learning}} \\
\quad TREC
& 65.50 & 43.50 & \textbf{66.00} & \textbf{66.00} & 57.00 & 62.00 \\
\quad TriviaQA
& 87.05 & 84.48 & 87.50 & 87.03 & 82.82 & \textbf{87.52} \\
\quad SAMSum
& 34.91 & 31.07 & 35.76 & 34.35 & 29.74 & \textbf{36.20} \\
\quad LSHT
& 39.98 & 19.50 & 35.56 & 36.85 & 31.10 & \textbf{37.00} \\

\addlinespace[2pt]
\multicolumn{7}{l}{\textit{Synthetic tasks}} \\
\quad PassageCount
& 5.00 & 3.00 & 5.00 & \textbf{8.00} & 2.17 & 6.00 \\
\quad PassageRetrieval-en
& 99.00 & 14.00 & 94.50 & 86.00 & 61.00 & \textbf{97.50} \\
\quad PassageRetrieval-zh
& 82.50 & 22.00 & 77.00 & 75.50 & 26.00 & \textbf{83.00} \\

\addlinespace[2pt]
\multicolumn{7}{l}{\textit{Code completion}} \\
\quad LCC
& 56.79 & 54.84 & 59.50 & \textbf{60.11} & 29.37 & 58.32 \\
\quad RepoBench-P
& 63.92 & 55.64 & 62.87 & \textbf{64.39} & 59.42 & 63.38 \\

\midrule
\textbf{Avg.}
& 44.79 & 27.29 & 43.28 & 42.04 & 32.68 & \textbf{44.00} \\
\bottomrule
\end{tabular*}
\end{table}

\subsection{Task-Level RULER Results at 128K}
\label{app:ruler-results}

Tables~\ref{tab:ruler-128k-llama31} and
\ref{tab:ruler-128k-qwen25} report the complete results on all 13 RULER
tasks at a context length of 128K. The group averages correspond to the
eight task groups reported in Figure~\ref{fig:ruler}.

\begin{table}[h]
\centering
\caption{Task-level RULER results at 128K on Llama-3.1-8B-Instruct. Task Avg. averages all 13 tasks, while Group Avg. averages the eight task groups used in the main-text evaluation. Bold indicates the best KV cache compression method in each row.}
\label{tab:ruler-128k-llama31}
\setlength{\tabcolsep}{2.8pt}
\renewcommand{\arraystretch}{0.96}
\begin{tabular*}{\linewidth}{@{\extracolsep{\fill}}lcccccc@{}}
\toprule
\multirow{2}{*}{\textbf{Task}}
& \multirow{2}{*}{\textbf{Dense}}
& \textbf{Streaming}
& \textbf{Duo}
& \multirow{2}{*}{\textbf{MoA}}
& \multirow{2}{*}{\textbf{KVzip-CI}}
& \textbf{Distance} \\
& & \textbf{LLM} & \textbf{Attention} & & & \textbf{KV} \\
\midrule

\multicolumn{7}{l}{\textit{Single NIAH}} \\
\quad Single 1
& 100.00 & 0.80 & 99.60 & 51.80 & 0.00 & \textbf{100.00} \\
\quad Single 2
& 98.60 & 2.20 & \textbf{97.60} & 72.20 & 0.00 & 95.60 \\
\quad Single 3
& 99.60 & 4.40 & 97.80 & 51.40 & 0.00 & \textbf{99.40} \\

\addlinespace[2pt]
\multicolumn{7}{l}{\textit{Multi-key NIAH}} \\
\quad Multi-key 1
& 95.20 & 4.00 & 95.00 & 68.80 & 0.00 & \textbf{95.60} \\
\quad Multi-key 2
& 73.40 & 0.40 & 54.20 & 39.60 & 0.00 & \textbf{64.60} \\
\quad Multi-key 3
& 58.80 & 0.20 & 37.60 & 18.80 & 0.00 & \textbf{40.80} \\

\addlinespace[2pt]
\multicolumn{7}{l}{\textit{Other retrieval tasks}} \\
\quad Multi-value
& 93.35 & 4.80 & 88.80 & 49.20 & 0.00 & \textbf{91.05} \\
\quad Multi-query
& 97.65 & 3.20 & 94.25 & 65.50 & 0.00 & \textbf{97.25} \\

\addlinespace[2pt]
\multicolumn{7}{l}{\textit{Tracking and aggregation}} \\
\quad Variable tracking
& 56.76 & 1.24 & 46.80 & 55.00 & 0.00 & \textbf{70.12} \\
\quad Common words
& 0.04 & \textbf{0.28} & 0.08 & 0.10 & 0.00 & 0.14 \\
\quad Frequent words
& 73.87 & 68.07 & 65.60 & \textbf{76.47} & 2.73 & 74.00 \\

\addlinespace[2pt]
\multicolumn{7}{l}{\textit{Question answering}} \\
\quad QA 1
& 71.40 & 14.40 & 57.60 & 51.40 & 9.80 & \textbf{63.80} \\
\quad QA 2
& 43.60 & 18.60 & 37.60 & 30.60 & 7.40 & \textbf{39.00} \\

\midrule
\textbf{Task Avg.}
& 74.02 & 9.43 & 67.12 & 48.53 & 1.53 & \textbf{71.64} \\
\textbf{Group Avg.}
& 69.30 & 12.26 & 62.97 & 48.52 & 1.42 & \textbf{68.66} \\
\bottomrule
\end{tabular*}
\end{table}

\begin{table}[h]
\centering
\caption{Task-level RULER results at 128K on Qwen2.5-7B-Instruct. Task Avg. averages all 13 tasks, while Group Avg. averages the eight task groups used in the main-text evaluation. Bold indicates the best KV cache compression method in each row.}
\label{tab:ruler-128k-qwen25}
\setlength{\tabcolsep}{2.8pt}
\renewcommand{\arraystretch}{0.96}
\begin{tabular*}{\linewidth}{@{\extracolsep{\fill}}lcccccc@{}}
\toprule
\multirow{2}{*}{\textbf{Task}}
& \multirow{2}{*}{\textbf{Dense}}
& \textbf{Streaming}
& \textbf{Duo}
& \multirow{2}{*}{\textbf{MoA}}
& \multirow{2}{*}{\textbf{KVzip-CI}}
& \textbf{Distance} \\
& & \textbf{LLM} & \textbf{Attention} & & & \textbf{KV} \\
\midrule

\multicolumn{7}{l}{\textit{Single NIAH}} \\
\quad Single 1
& 89.80 & 0.80 & \textbf{100.00} & 75.80
& \textbf{100.00} & \textbf{100.00} \\
\quad Single 2
& 27.00 & 3.00 & 98.40 & 73.60 & 90.40 & \textbf{99.20} \\
\quad Single 3
& 20.00 & 3.00 & 91.80 & 74.20 & 85.60 & \textbf{96.60} \\

\addlinespace[2pt]
\multicolumn{7}{l}{\textit{Multi-key NIAH}} \\
\quad Multi-key 1
& 27.40 & 4.20 & 92.20 & 66.40 & 66.80 & \textbf{94.20} \\
\quad Multi-key 2
& 2.00 & 0.00 & 30.60 & 25.60 & 13.60 & \textbf{52.40} \\
\quad Multi-key 3
& 0.20 & 0.60 & 5.80 & 9.00 & 3.00 & \textbf{20.40} \\

\addlinespace[2pt]
\multicolumn{7}{l}{\textit{Other retrieval tasks}} \\
\quad Multi-value
& 24.25 & 4.75 & 54.45 & 56.05 & 57.75 & \textbf{71.60} \\
\quad Multi-query
& 16.25 & 4.20 & 78.65 & 66.35 & 73.10 & \textbf{90.05} \\

\addlinespace[2pt]
\multicolumn{7}{l}{\textit{Tracking and aggregation}} \\
\quad Variable tracking
& 6.20 & 0.80 & 65.96 & 66.88 & 71.88 & \textbf{81.92} \\
\quad Common words
& 4.60 & 0.90 & 34.16 & 28.74 & 19.48 & \textbf{40.90} \\
\quad Frequent words
& 58.80 & 62.80 & \textbf{67.33} & 49.80 & 57.00 & 64.87 \\

\addlinespace[2pt]
\multicolumn{7}{l}{\textit{Question answering}} \\
\quad QA 1
& 22.80 & 18.00 & 38.60 & \textbf{71.20} & 29.80 & 49.20 \\
\quad QA 2
& 21.60 & 18.00 & 26.40 & 34.60 & 24.20 & \textbf{37.60} \\

\midrule
\textbf{Task Avg.}
& 24.68 & 9.31 & 60.33 & 53.71 & 53.28 & \textbf{69.15} \\
\textbf{Group Avg.}
& 23.47 & 11.91 & 59.08 & 53.61 & 53.25 & \textbf{68.38} \\
\bottomrule
\end{tabular*}
\end{table}

\section{Additional Efficiency Results}
\label{app:additional-efficiency}

\begin{figure}[h]
    \centering
    \includegraphics[width=\linewidth]{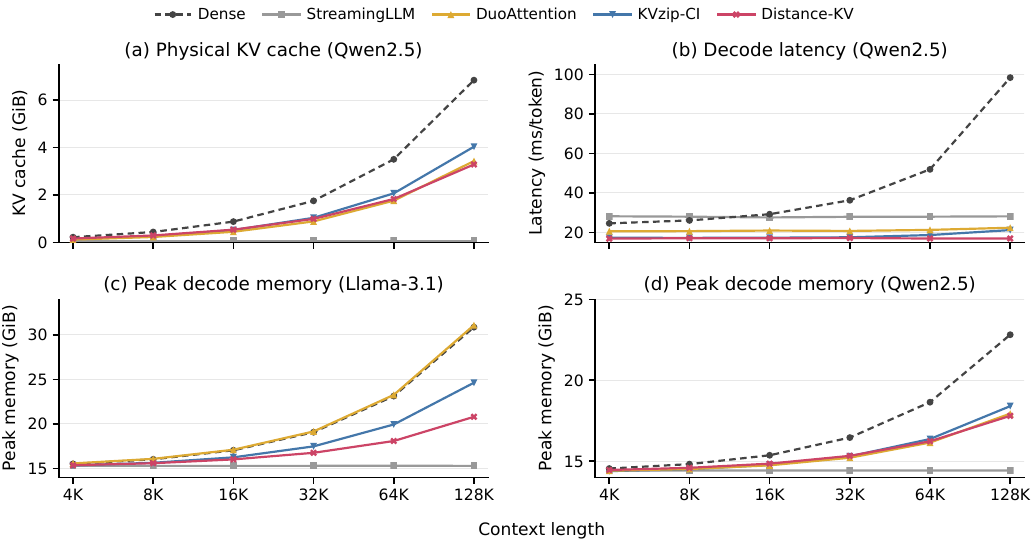}
    \caption{Additional efficiency results on Llama-3.1-8B-Instruct and Qwen2.5-7B-Instruct. (a)--(b) Physical KV cache size and per-token decoding latency on Qwen2.5-7B-Instruct. (c)--(d) Peak allocated GPU memory during decoding on both models. All measurements use a single H100 GPU with a batch size of 1.}
    \label{fig:additional-efficiency}
\end{figure}

Figure~\ref{fig:additional-efficiency}(a)--(b) shows that, on Qwen2.5-7B-Instruct, the physical KV cache of Distance-KV grows substantially more slowly with context length than that of Dense, while maintaining the lowest decoding latency among the evaluated methods. At 128K, Distance-KV uses 3.28 GiB of physical KV cache, a 52.0\% reduction from Dense, and achieves a decoding latency of 16.96 ms/token, corresponding to a \(5.80\times\) speedup. Figure~\ref{fig:additional-efficiency}(c)--(d) further shows that physical cache compression reduces overall memory usage during decoding. At 128K, Distance-KV uses 20.80 GiB and 17.80 GiB of peak allocated memory on Llama-3.1-8B-Instruct and Qwen2.5-7B-Instruct, reducing memory usage relative to Dense by 32.6\% and 22.0\%, respectively. These results demonstrate consistent reductions in KV cache size, decoding latency, and runtime memory across backbone models.

\section{Additional Ablation Results}
\label{app:additional-ablation}

\begin{table}[h]
\centering
\caption{Task-level structural ablation results on RULER at a context length of 128K using Llama-3.1-8B-Instruct. All variants are matched in physical KV retention, and Avg. denotes the mean over all 13 tasks. Bold indicates the best result in each row.}
\label{tab:ruler-ablation-task-level}
\vspace{3pt}
\setlength{\tabcolsep}{6pt}
\begin{tabular*}{\linewidth}{@{\extracolsep{\fill}}lccccc@{}}
\toprule
\multicolumn{1}{c}{\multirow{2}{*}{Task}}
& Head- & Distance- & Distance- & Fully & Distance- \\
& only & only & shuffled & shuffled & KV \\
\midrule

\multicolumn{6}{l}{\textit{Single NIAH}} \\
\quad Single 1
& 81.00 & 14.80 & 74.00 & 19.00 & \textbf{93.60} \\
\quad Single 2
& 73.20 & 10.00 & 87.20 & 13.00 & \textbf{95.60} \\
\quad Single 3
& 67.20 & 5.20 & 93.00 & 7.00 & \textbf{98.60} \\

\addlinespace[2pt]
\multicolumn{6}{l}{\textit{Multi-key NIAH}} \\
\quad Multi-key 1
& 69.60 & 12.80 & 87.40 & 14.80 & \textbf{92.60} \\
\quad Multi-key 2
& 6.60 & 0.20 & 22.20 & 4.80 & \textbf{36.80} \\
\quad Multi-key 3
& 0.20 & 0.20 & 1.20 & 1.20 & \textbf{7.00} \\

\addlinespace[2pt]
\multicolumn{6}{l}{\textit{Other retrieval tasks}} \\
\quad Multi-value
& 84.85 & 16.30 & 90.10 & 9.45 & \textbf{93.25} \\
\quad Multi-query
& 87.25 & 13.80 & 92.35 & 11.70 & \textbf{95.50} \\

\addlinespace[2pt]
\multicolumn{6}{l}{\textit{Tracking and aggregation}} \\
\quad Variable tracking
& 2.56 & 3.56 & 33.60 & 9.28 & \textbf{61.36} \\
\quad Common words
& 0.02 & 0.08 & 0.14 & 0.04 & \textbf{0.18} \\
\quad Frequent words
& 59.93 & 55.80 & 64.53 & \textbf{76.60} & 70.33 \\

\addlinespace[2pt]
\multicolumn{6}{l}{\textit{Question answering}} \\
\quad QA 1
& 27.20 & 15.60 & 41.60 & 35.40 & \textbf{53.00} \\
\quad QA 2
& 18.60 & 16.40 & 24.00 & 27.40 & \textbf{30.60} \\

\midrule
Avg.
& 44.48 & 12.67 & 54.72 & 17.67 & \textbf{63.72} \\
\bottomrule
\end{tabular*}
\end{table}

Table~\ref{tab:ruler-ablation-task-level} reports the complete task-level results for the structural ablations and additionally includes the fully shuffled variant. This variant disrupts all correspondences among layers, query heads, and relative-distance bins while preserving the logical query-head and physical KV retention ratios. Fully shuffled reduces the overall average from 63.72 for Distance-KV to 17.67 and causes substantial degradation on all eight retrieval tasks. Although it achieves a higher score on Frequent words, this isolated improvement does not extend to other tasks or overall performance. Together with the Distance-shuffled results, these findings show that matching the cache size alone is insufficient and that the learned layer--query-head--distance structure is critical to preserving long-context capabilities.

\end{document}